\documentclass[letterpaper, 10 pt, journal, twoside]{IEEEtran}
\ifCLASSINFOpdf
\else
\fi

\usepackage[utf8]{inputenc}
\usepackage[T1]{fontenc}
\usepackage{cite}
\usepackage{amsmath,amssymb,amsfonts}
\usepackage{algorithmic}
\usepackage{graphicx}
\usepackage{textcomp}
\usepackage{authblk}
\usepackage{gensymb}

\usepackage{hyperref}
\usepackage{scrextend}
\usepackage{changepage}
\usepackage[margin=0.75in]{geometry}
\usepackage{multirow}
\usepackage{amsfonts}
\usepackage[table]{xcolor}
\usepackage{booktabs}
\usepackage{mathtools}
\DeclarePairedDelimiter\ceil{\lceil}{\rceil}
\DeclarePairedDelimiter\floor{\lfloor}{\rfloor}
\DeclareMathOperator*{\argmax}{argmax}

\begin{document}
%
% paper title
% Titles are generally capitalized except for words such as a, an, and, as,
% at, but, by, for, in, nor, of, on, or, the, to and up, which are usually
% not capitalized unless they are the first or last word of the title.
% Linebreaks \\ can be used within to get better formatting as desired.
% Do not put math or special symbols in the title.
\title{Can I Pour into It? Robot Imagining Open Containability Affordance of Previously Unseen Objects via Physical Simulations}
%
%
% author names and IEEE memberships
% note positions of commas and nonbreaking spaces ( ~ ) LaTeX will not break
% a structure at a ~ so this keeps an author's name from being broken across
% two lines.
% use \thanks{} to gain access to the first footnote area
% a separate \thanks must be used for each paragraph as LaTeX2e's \thanks
% was not built to handle multiple paragraphs
%

\author{Hongtao Wu$^{1}$ and Gregory S. Chirikjian$^{1, 2}$%
\thanks{Manuscript received August 15, 2020; accepted October 16, 2020.}%Use only for final RAL version
\thanks{This paper was recommended for publication by Editor Nancy Amato upon evaluation of the Associate Editor and Reviewers' comments.
This work was supported by Office of Naval Research Award N00014-17-1-2142 and National Science Foundation grant IIS-1619050.} %Use only for final RAL version
\thanks{$^{1}$Hongtao Wu and Gregory S. Chirikjian are with the Laboratory for Computational Sensing and Robotics, the Johns Hopkins University, Baltimore, MD 21218 USA {\tt\footnotesize \{hwu67, gchirik1\}@jhu.edu}
        }%

\thanks{$^{2} $Gregory S. Chirikjian is also with the Department of Mechanical Engineering, National University of Singapore, Singapore 117575 {\tt\footnotesize mpegre@nus.edu.sg}
        }%
\thanks{Digital Object Identifier (DOI): see top of this page.}
}
% note the % following the last \IEEEmembership and also \thanks - 
% these prevent an unwanted space from occurring between the last author name
% and the end of the author line. i.e., if you had this:
% 
% \author{....lastname \thanks{...} \thanks{...} }
%                     ^------------^------------^----Do not want these spaces!
%
% a space would be appended to the last name and could cause every name on that
% line to be shifted left slightly. This is one of those "LaTeX things". For
% instance, "\textbf{A} \textbf{B}" will typeset as "A B" not "AB". To get
% "AB" then you have to do: "\textbf{A}\textbf{B}"
% \thanks is no different in this regard, so shield the last } of each \thanks
% that ends a line with a % and do not let a space in before the next \thanks.
% Spaces after \IEEEmembership other than the last one are OK (and needed) as
% you are supposed to have spaces between the names. For what it is worth,
% this is a minor point as most people would not even notice if the said evil
% space somehow managed to creep in.

% The paper headers
\markboth{IEEE Robotics and Automation Letters. Preprint Version. October, 2020}
{Wu \MakeLowercase{\textit{et al.}}: Robot Imagination of Open Containability Affordance} 

% The only time the second header will appear is for the odd numbered pages
% after the title page when using the twoside option.
% 
% *** Note that you probably will NOT want to include the author's ***
% *** name in the headers of peer review papers.                   ***
% You can use \ifCLASSOPTIONpeerreview for conditional compilation here if
% you desire.

% If you want to put a publisher's ID mark on the page you can do it like
% this:
%\IEEEpubid{0000--0000/00\$00.00~\copyright~2015 IEEE}
% Remember, if you use this you must call \IEEEpubidadjcol in the second
% column for its text to clear the IEEEpubid mark.

% use for special paper notices
%\IEEEspecialpapernotice{(Invited Paper)}

% make the title area
\maketitle

% As a general rule, do not put math, special symbols or citations
% in the abstract or keywords.
\begin{abstract}
Open containers, \textit{i.e.}, containers without covers, are an important and ubiquitous class of objects in human life.
In this letter, we propose a novel method for robots to ``imagine'' the open containability affordance of a previously unseen object via physical simulations. 
The robot autonomously scans the object with an RGB-D camera.
The scanned 3D model is used for open containability imagination which quantifies the open containability affordance by physically simulating dropping particles onto the object and counting how many particles are retained in it.
This quantification is used for open-container vs. non-open-container binary classification (hereafter referred to as open container classification).
If the object is classified as an open container, the robot further imagines pouring into the object, again using physical simulations, to obtain the pouring position and orientation for real robot autonomous pouring. 
We evaluate our method on open container classification and autonomous pouring of granular material on a dataset containing 130 previously unseen objects with 57 object categories.
Although our proposed method uses only 11 objects for simulation calibration, its open container classification aligns well with human judgements.
In addition, our method endows the robot with the capability to autonomously pour into the 55 containers in the dataset with a very high success rate.
We also compare to a deep learning method.
Results show that our method achieves the same performance as the deep learning method on open container classification and outperforms it on autonomous pouring.
Moreover, our method is fully explainable.
\end{abstract}

% Note that keywords are not normally used for peerreview papers.
% \begin{IEEEkeywords}
% IEEE, IEEEtran, journal, \LaTeX, paper, template.
% \end{IEEEkeywords}

% For peer review papers, you can put extra information on the cover
% page as needed:
% \ifCLASSOPTIONpeerreview
% \begin{center} \bfseries EDICS Category: 3-BBND \end{center}
% \fi
%
% For peerreview papers, this IEEEtran command inserts a page break and
% creates the second title. It will be ignored for other modes.
\IEEEpeerreviewmaketitle

\section{Introduction}
\label{sec:1}

\begin{figure}
    \centering
    \includegraphics[scale=0.35]{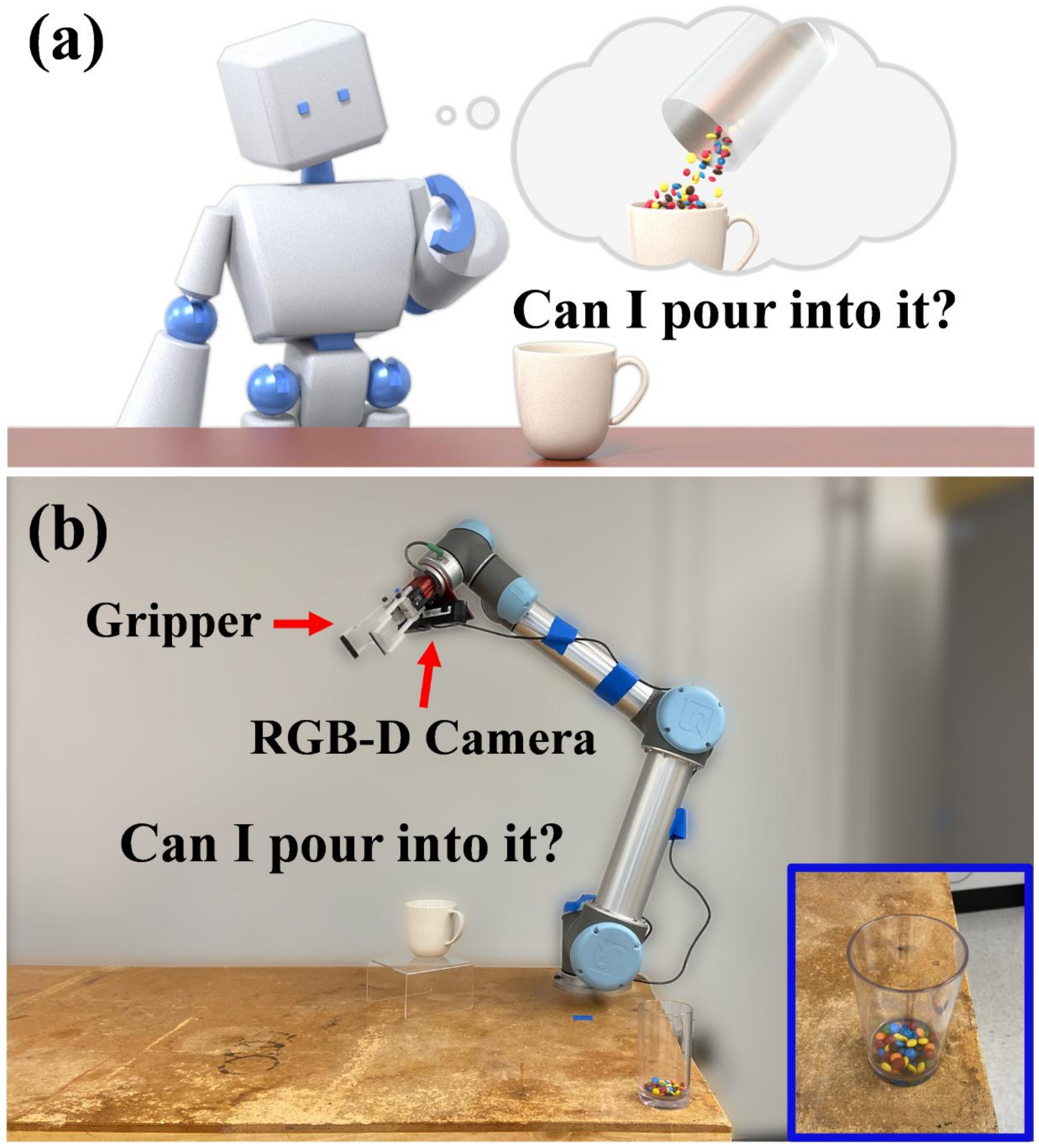}
    \caption{\textbf{Robot Imagination.} (a) The robot imagines the open containability affordance of an object by physically simulating potential interactions with the object. 
    (b) Real-world experiment setup of robot imagination.
    }
    \label{fig:1}
    \vspace{-0.6cm}
\end{figure}

\begin{figure*}
    \centering
    \includegraphics[scale=0.47]{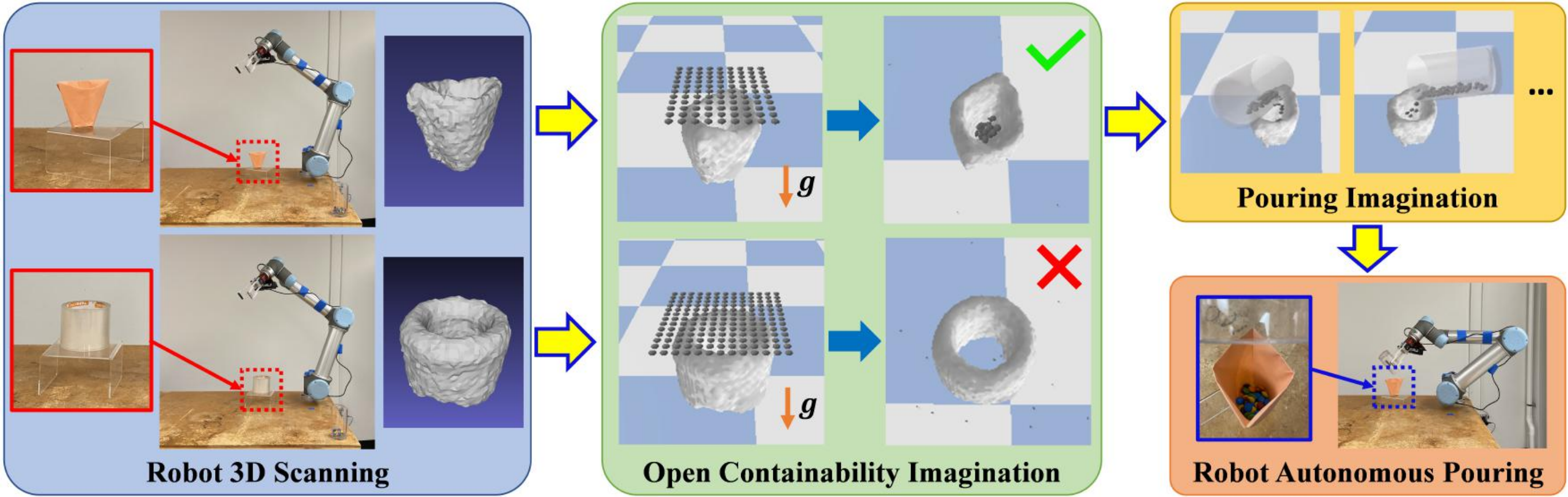}
    \caption{\textbf{Pipeline.} 
    The object of interest is randomly placed on a transparent platform for scanning. 
    The two objects shown in the figure are an origami cup and a roll of tape. 
    First, the robot autonomously scans the object to obtain the 3D model of the object. 
    The robot uses the 3D model for open containability imagination and pouring imagination.
    Between the two objects, only the origami cup is identified as an open container (checkmark in the middle figure).}
    \label{fig:2}
    \vspace{-0.5cm}
\end{figure*}

\IEEEPARstart{T}{he} understanding of object affordances \cite{gibson1979ecological}, \textit{i.e.}, the set of possible interactions with an object, plays a crucial role in human perception of real-world objects despite the tremendous variance in their appearances \cite{dicarlo2012does}. 
As robots begin to enter our daily life, it is necessary for them to assess previously unseen real-world objects and understand how they can be interacted with via affordance-based object perception.
Despite the necessity and advantages of affordance reasoning in object perception, the majority of work on this topic is appearance-based. 
Appearance-based methods can be brittle when encountering objects with large intra-class variations.
Moreover, they can be challenged when functional reasoning about previously unseen objects and inter-class function generalization are needed.
For example, consider a robot trying to prepare a bowl of M\&M's\textsuperscript{\textregistered} candies (hereafter referred to as candies) in a kitchen. 
What if the bowl in the kitchen has a different appearance from the bowls that the robot has seen before? 
What if the bowl is broken with a hole at the bottom while still visually appears to be a bowl?
If there are no bowls but only cups in the kitchen, can the robot be programmed to \textit{improvise}? 
Given these unexpected situations, we argue that appearance-based object perception alone is not sufficient for robotics application in the real world; what is missing is the capability to ``physically understand'' object affordances.
If the bowl is previously unseen, the robot should recognize its containability affordance; if the bowl is broken, it should reason that the bowl cannot afford to contain candies; if there are no bowls but only cups, it should generalize the containability affordance from bowls to cups and use a cup instead to contain the candies (Fig. \ref{fig:1}).

To address the problem of affordance reasoning in object perception, we introduced the \textit{interaction-based definition} of an object and a method to ``imagine'' object affordances as a chair via physical simulations in our previous work \cite{wu2020chair}. 
In this paper, we extend our idea of physical simulation-based object affordance reasoning to the class of open containers and propose a novel method for robots to imagine the open containability affordance.  
We are inspired by the fact that humans mentally simulate high-level physical interactions to reason about complex systems \cite{battaglia2013simulation}. 
Our long-term goal is to endow robots with similar mental capabilities to imagine potential interactions with an object and reason about its object affordances accordingly (Fig. \ref{fig:1}(a)). 

We first discuss the physical attributes to imagine for the class of open containers.
Containers can be generally categorized into open containers (\textit{e.g.}, cups) and closed containers (\textit{e.g.}, bottles with a cap).
The \textit{containability closure} \cite{varadarajan2012afnet} of a closed container makes the physical interaction with it different from that with an open container.
An agent needs to open a closed container to transfer contents into it while there is no such need for an open container.
The open containability we describe in this paper does not take into account the containability closure.
The pose of an open container also affects the interaction with it.
An upside down open container needs to be reoriented upright to afford open containability. 
Finally, containability is related to the physical properties of the container and the contents. 
For example, the origami cup in Fig. \ref{fig:2} cannot contain water because it will be soaked but it is able to contain candies. 
In this paper, we focus on reasoning the open containability for granular material of an upright object. 
We propose the \textit{interaction-based definition} of open containers:
\begin{adjustwidth}{0.50cm}{0.50cm}
``an object which can be placed such that liquid or granular material can be poured into it from above and retained therein under small perturbations in pose.''
\end{adjustwidth}

The interaction-based definition defines open containers from the perspective of object interaction which robots can ``understand'' and ``imagine'' accordingly.
Fig. \ref{fig:2} shows our method's pipeline.
In our method, the robot performs open containability imagination (Sec. \ref{sec:3A}) to quantify the open containability affordance of a previously unseen object.
This quantification is used as a cue for open-container vs. non-open-container binary classification (hereafter referred to as open container classification). 
If the object is classified as an open container, the robot further performs pouring imagination (Sec. \ref{sec:3B}) to obtain the pouring position and orientation for the robot to pour autonomously in the real world (Sec. \ref{sec:4D}).
We implement our imagination method on a UR5 manipulator.
We evaluate our method on open container classification and autonomous pouring of granular material (M\&M's\textsuperscript{\textregistered} candies) with a dataset containing 130 previously unseen real-world objects. 
The dataset has 57 object categories. 
Results show that our method is able to achieve human-like judgements on open container classification, even though it uses only 11 objects for simulation calibration. 
In addition, our method is able to autonomously pour into the 55 containers in the dataset with a very high success rate. 
We also compare to a deep learning method \cite{do2018affordancenet} which we find the most relevant in the literature.
Our method is able to achieve the same performance as the deep learning method on open container classification and outperform it on autonomous pouring.
More details can be found on our project page: \url{https://chirikjianlab.github.io/realcontainerimagination/}

\section{Related Work}
\textbf{Containability Reasoning.}
Containability reasoning has been studied in the field of cognitive science \cite{liang2015evaluating, ullman2019model} and computer vision \cite{varadarajan2012afnet, yu2015fill, hinkle2013predicting, liang2018tracking}. 
Liang \textit{et al.} \cite{liang2015evaluating} compare human judgements of containing relations between different objects with physical simulation results. Yu \textit{et al.} \cite{yu2015fill} voxelize the world containing an object of interest and determine the state of each voxel to reason about whether the object is a container and its best filling direction.
Our method diverges from these approaches by reasoning open containability affordance with a real robot and leveraging the affordance reasoning on robot autonomous pouring.

\textbf{Learning-based Affordance Detection.}
There is a growing interest in learning object affordances and object functionalities in the computer vision and robotics community \cite{myers2015affordance, do2018affordancenet, nguyen2017object, chu2019learning, sawatzky2017weakly, roy2016multi, manuelli2019kpam, ruiz2020geometric, zhu2014reasoning, zhu2018visual}. 
\cite{nguyen2017object, do2018affordancenet, chu2019learning, roy2016multi, sawatzky2017weakly} leverage convolutional neural networks (CNNs) to segment affordance regions within an image. 
Manuelli \textit{et al.} \cite{manuelli2019kpam} use deep learning to identify keypoint affordances for objects to reinforce robot manipulation.
Instead of learning appearance-based cues, our method digs into the underlying physics by using physical simulations to encode object affordances.

\textbf{Physics-based Reasoning.}
The concept of object affordances has also been studied by introducing physics to reason about the functionality of an object \cite{zhu2018visual, ho1987representing, wu2020chair, yu2015fill, hinkle2013predicting, kunze2017envisioning, abelha2017learning, boschert2016digital}.
Kunze and Beetz \cite{kunze2017envisioning} reason the outcome of robot actions with simulation-based projections to guide robots in planning actions.
The concept of digital twins \cite{boschert2016digital} uses simulations to replicate physical systems for operation optimization and failure prediction in manufacturing.
Instead of reasoning physics or using physics to predict action outcomes, we exploit physics to imagine potential interactions with an object for object classification and robot-object interaction.

\textbf{Autonomous Pouring.}
Autonomous pouring is an important task in robot manipulation \cite{guevara2017adaptable, yamaguchi2015differential, schenck2017visual, pan2016motion, kennedy2019autonomous}.
Yamaguchi and Atkeson \cite{yamaguchi2015differential} explore temporally decomposed dynamics in pouring simulation experiments with differential dynamic programming. 
Scheneck and Fox \cite{schenck2017visual} learn to estimate the liquid volume in a container and incorporate this sensory feedback into robotic pouring.
Unlike these methods which know the object class (container) and/or the object 3D model \textit{a priori}, our method categorizes a previously unseen object and predict the pouring position and orientation without knowing the 3D model or the category of the object \textit{a priori}.

\section{Method} 
Given the scanned 3D model of the object, our method attaches the body frame to its center of mass (CoM) and set the axes of the frame parallel to those of the world frame. 
The object and particles are considered as rigid bodies. 
Rigid body transformation can be specified by $g=(R, \textbf{t}) \in SE(3)$ where $R \in SO(3)$ is the rotation matrix and $\textbf{t}=[x, y, z]^{T} \in \mathbb{R}^{3}$ parameterizes the translation.
The ground in the imagination is set as a horizontal plane below and not in contact with the object.

\subsection{Open Containability Imagination}
\label{sec:3A}
\begin{figure}
    \centering
    \includegraphics[width=1\columnwidth]{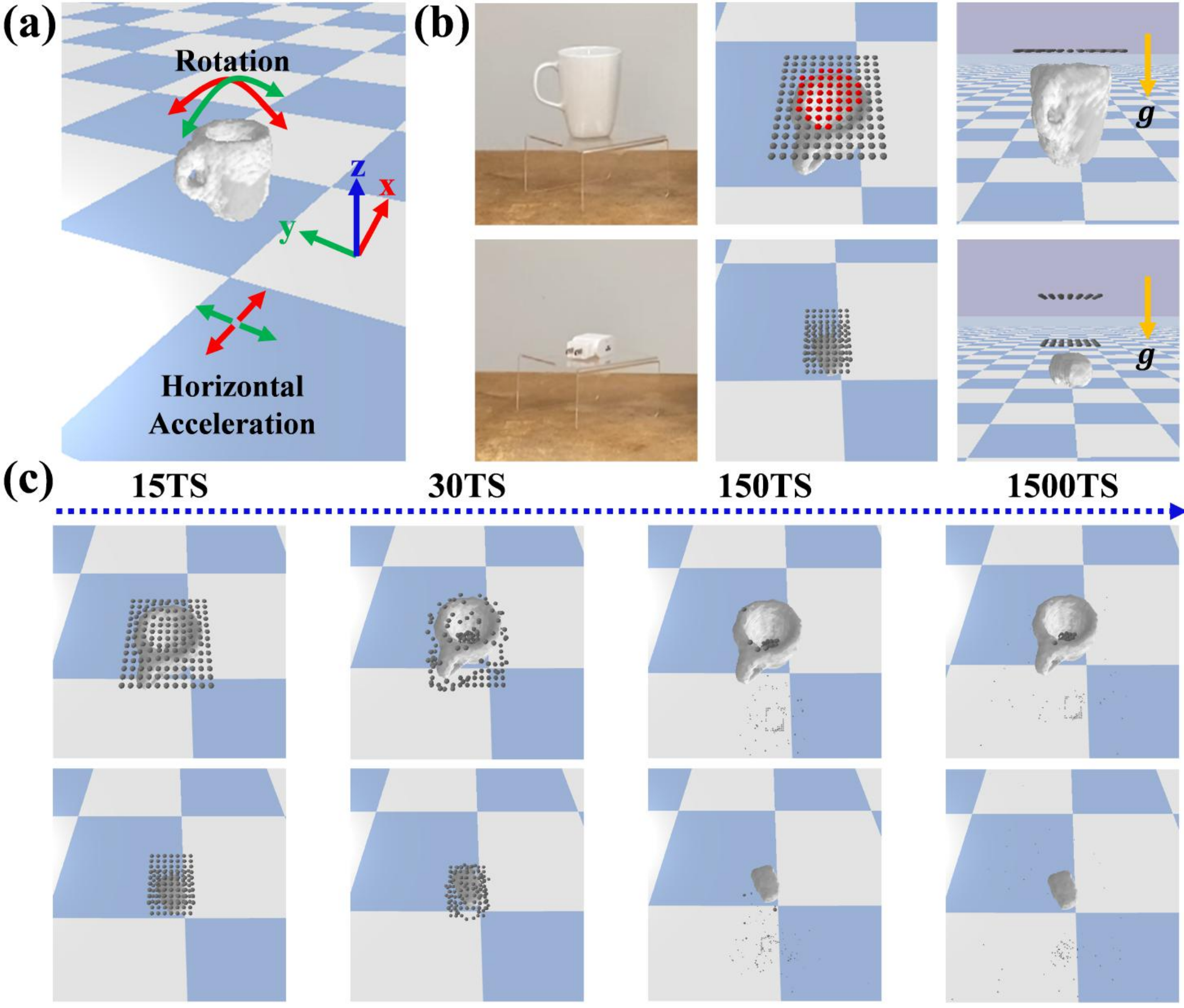}
    \caption{\textbf{Open Containability Imagination.}
    (a) Perturbation setting. The open containability imagination lasts for $T_O=1500$ timesteps (1/240 seconds per timestep in world clock). 
    (b) The particle dropping pattern of a mug and a USB charger. 
    The highlighted red particles in the dropping pattern of the mug show the \textit{footprint} of the mug.
    Note that $N_z=2$ for the USB charger. 
    The lowest layer is 1cm above the object AABB; the distance between two adjacent layers is 5cm. 
    (c) Open containability imagination of the two objects in (b) at different timesteps (TS).}
    \label{fig:3}
    \vspace{-0.5cm}
\end{figure}

Fig. \ref{fig:3} shows the setting of open containability imagination.
Particles are placed above the axis-aligned bounding box (AABB) of the object and patterned in a grid to cover the top face of the AABB (Fig. \ref{fig:3}(b)).
We denote the numbers of particles along the x-, y-, and z-axes as  $N_{x}, N_{y}, N_{z} \in \mathbb{Z}^{+}$. $N_{x}$, $N_{y}$ are set as follows:
\begin{equation*}
    N_{x}(s) = \floor*{\frac{sl^{x}_{o}}{l^{x}_{p}}}, 
    N_{y}(s) = \floor*{\frac{sl^{y}_{o}}{l^{y}_{p}}}
\end{equation*}
where $l^{x}_{o}$ ($l^{x}_{p}$) and $l^{y}_{o}$ ($l^{y}_{p}$) are the extents of the object (particle) AABB along the x- and y-axes, respectively; $s$ is a scale factor to avoid cases where there are too many particles to drop which will greatly increase the runtime; $\floor*{\cdot}$ is the floor function.
We first set $s=1, N_z=1$ and calculate $N_x(1)$ and $N_y(1)$. 
If $N_x(1)N_y(1)>N_{max}$, \textit{i.e.}, there are too many particles, we decrease $s$ by setting $s=\sqrt{\frac{N_{max}}{N_x(1)N_y(1)}}$ and recalculate $N_x(s)$ and $N_y(s)$. 
If $N_x(1)N_y(1)<N_{min}$, \textit{i.e.}, there are too few particles which will make later calculation of open containability (Eqn. \ref{eqn:1}) unreliable, we increase $N_z$ by setting $N_z=\ceil*{\frac{N_{min}}{N_x(1)N_y(1)}}$.
$\ceil*{\cdot}$ is the ceiling function.
$N_{max}$ and $N_{min}$ are two thresholds.

After the particles are set up, they are dropped in the physical simulation with gravity.
According to the interaction-based definition, if an object can robustly retain particles, small perturbations in pose shall not spill the particles.
Therefore, perturbations are introduced by simulating motions equivalent to rotating and translating the object when the drop is finished (Fig. \ref{fig:3}(a)).
The object is rotated by a small degree ($\pm\pi/60$) along the x- and y-axes to simulate rotation; horizontal force fields ($g_{h}=0.5N/kg$) along the positive and negative directions of x- and y-axes are applied sequentially to simulate translation.
Note that applying a horizontal force field is equivalent to accelerating the object horizontally along the negative direction of the force field.
Translation is simulated with accelerations instead of constant velocities because based on common experience, spillage mostly happens during accelerations instead of moving with constant velocities. 
The number of particles retained within the objects, $N_{in}$, is obtained by counting the number of particles within the object AABB at the end of the simulation.
The open containability of the object $\Omega$ is quantified as:
\begin{equation}
    \Omega = \frac{N_{in}}{N_{drop}}
    \label{eqn:1}
\end{equation}
where $N_{drop} = N_xN_yN_z$ is the number of particles dropped. 
If $\Omega > \Omega_{thr}$, the object is considered as an open container. $\Omega_{thr}$ is a threshold.
The xy-plane projection of the initial dropping position of the particles retained within the object after the drop constitutes what we called the \textit{footprint} of the object (indicated in red in Fig. \ref{fig:3}(b)).

\begin{figure}
    \centering
    \includegraphics[width=1\columnwidth]{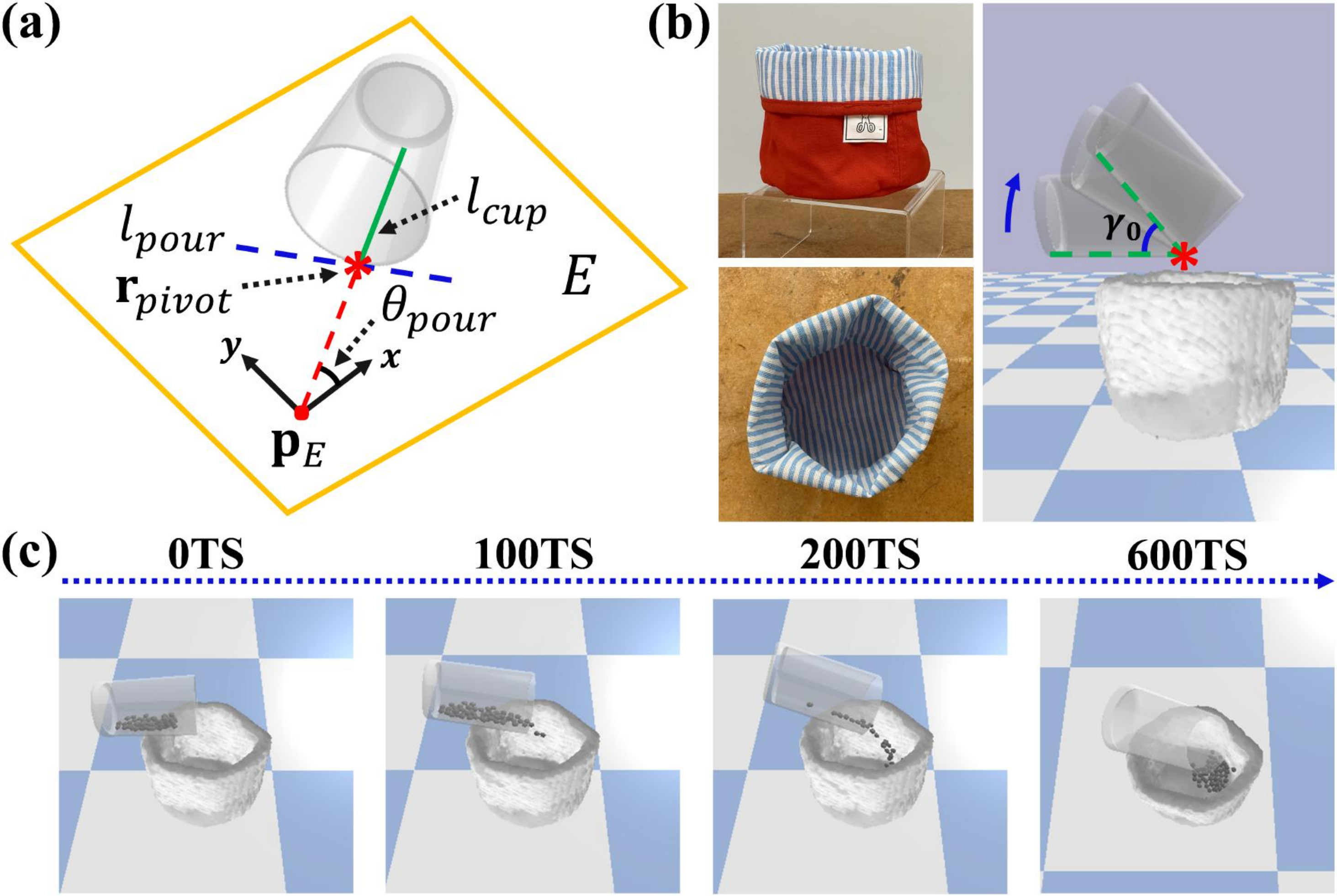}
    \caption{\textbf{Pouring Imagination.} 
    Each pouring lasts for $T_P=600$ timesteps in the simulation.
    (a) Pouring setting. 
    (b) Pouring imagination of a red bag. 
    We set $\gamma_{0} = 62\degree$ which is a hyperparameter specifying the total pouring angle. 
    (c) Snapshots of the pouring imagination at different timesteps (TS). 
    }
    \label{fig:4}
    \vspace{-0.5cm}
\end{figure}

\subsection{Pouring Imagination}
\label{sec:3B}
Pouring imagination simulates pouring particles from a cup from different positions and orientations (Fig. \ref{fig:4}).
To describe the pouring setting, we first define $\textbf{r}_{pivot} \in \mathbb{R}^{3}$ (red star in Fig. \ref{fig:4}(a)) as a point on the rim of the cup.
$\textbf{r}_{pivot}$ is fixed as the rotation pivot during the pouring. 
Since the cup used for pouring is symmetric about its longitudinal axis, any points on the rim can be equivalently used as the pivot.
For all pourings, we set $\textbf{r}_{pivot}$ 1cm above the object AABB.
That is, $\textbf{r}_{pivot}$ lives on a horizontal plane $E = \{(x, y, z) | z = z_{E}\}$.
The origin $\textbf{p}_{E} \in \mathbb{R}^2$ of $E$ is specified by the CoM of the object's footprint.
The x- and y-axes of $E$ are the corresponding principal axes.

The inner surface of the cup is a truncated cone.
We denote the slant height passing through $\textbf{r}_{pivot}$ as $l_{cup}$ (green line in Fig. \ref{fig:4}(a)).
For all pourings, the cup is initially placed such that $l_{cup} \in E$ and $\textbf{r}_{pivot}$ is the lowest end of the cup mouth.
This constrains the initial pose of the cup to $SE(2)$.
We parameterize a pouring with a tuple $(\theta_{pour}, \textbf{p}_{pour}) \in SE(2)$ where $\theta_{pour} \in S^{1}, \textbf{p}_{pour} \in \mathbb{R}^2$. 
At the beginning of each pouring, $l_{cup}$ is aligned with a ray (red dashed line Fig. \ref{fig:4}(a)) originated from $\textbf{p}_{E}$.
The angle between the x-axis of $E$ and the ray equals $\theta_{pour}$.
$\textbf{p}_{pour}$, which lies along the ray, specifies the $x$ and $y$ components of $\textbf{r}_{pivot}$ in $E$. 
The pouring is a one degree of freedom motion: the cup is rotated about an axis $l_{pour} \in E$ (blue dashed line in Fig. \ref{fig:4}(a)) which is tangent to the rim of the cup at $\textbf{r}_{pivot}$.

For an object, the imagination simulates pouring from 8 different $\theta_{pour}$:
\begin{equation}
    \theta_{pour}(i) = \frac{i\pi}{4}, i=0, 1, ..., 7
\end{equation}
For each $\theta_{pour}$, it simulates pouring at 3 different $\textbf{p}_{pour}$:
\begin{equation}
\begin{gathered}
    \textbf{p}_{pour}(i, j) = R(\theta_{pour}(i))
    \begin{bmatrix}
        L_{idt}(j)\\
        0
    \end{bmatrix}\\
\end{gathered}
\label{eqn:3}
\end{equation}
where $L_{idt}(j) = L_0 + j\Delta L, j = 0, 1, 2$ defines the j-th indentation distance; $L_0$ and $\Delta L$ are two hyperparameters: $L_0$=1cm and $\Delta L$ equals $1/3$ of the length of the diagonal of the object AABB's top face; $R(\theta) \in SO(2)$:
\begin{equation*}
    R = 
    \begin{bmatrix}
    \cos(\theta) & -\sin(\theta)\\
    \sin(\theta) & \cos(\theta)
    \end{bmatrix}
\end{equation*}

\begin{figure}
    \centering
    \includegraphics[width=1\columnwidth]{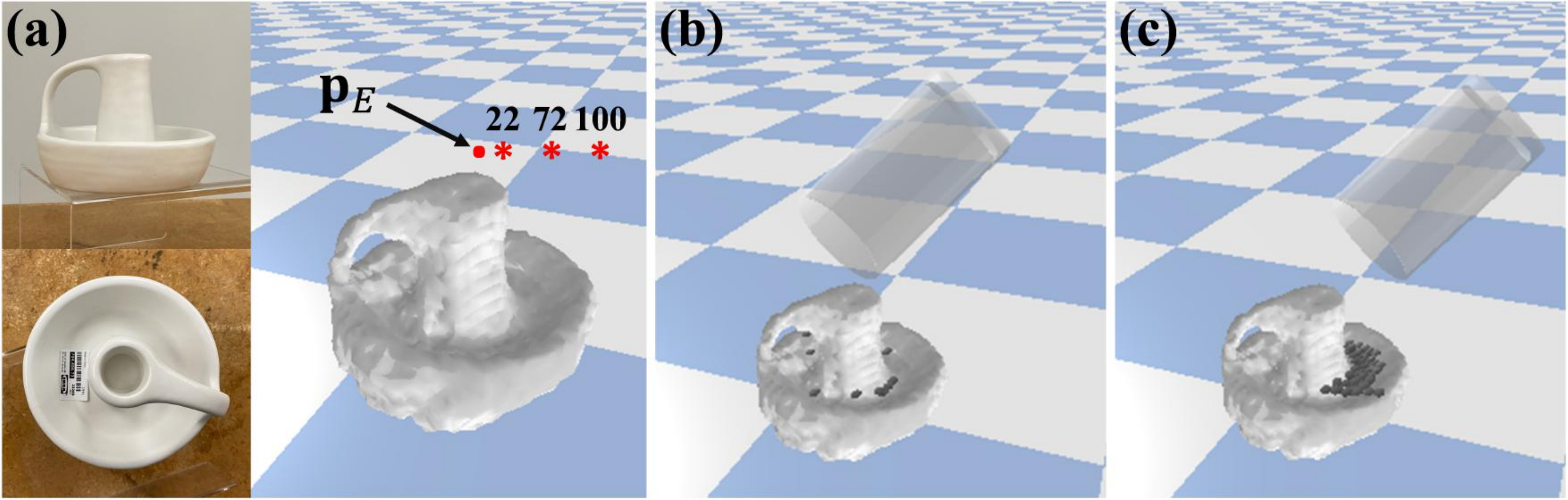}
    \caption{\textbf{Pouring Imagination of a Candlestick.} (a) The three red stars indicate $\textbf{p}_{pour}(6, 0)$, $\textbf{p}_{pour}(6, 1)$, and $\textbf{p}_{pour}(6, 2)$, respectively. The number above each red star are the corresponding particle-in-object ratio $P$ in percentage. The farther $\textbf{p}_{pour}$ is from $\textbf{p}_{E}$, the larger the value of $P$ is. (b) and (c) show the snapshot at the end of the pouring imagination pouring at $\textbf{p}_{pour}(6, 0)$ and $\textbf{p}_{pour}(6, 2)$, respectively.}
    \label{fig:5}
    \vspace{-0.5cm}
\end{figure}

The same method described in Sec. \ref{sec:3A} is used to count the number of particles retained within the object $N_{in}$ after each pouring. 
The particle-in-object ratio $P$ of pouring with $(\theta_{pour}, \textbf{p}_{pour})$ is specified as:
\begin{equation}
    P(\theta_{pour}, \textbf{p}_{pour}) =\frac{N_{in}}{N_{pour}}
    \label{eqn:pour}
\end{equation}
where $N_{pour}$ is the number of particles poured.
In total, the imagination simulates 24 pourings for 24 $(\theta_{pour}, \textbf{p}_{pour})$ tuples and get 24 $P$.
We add indentation $L_{idt}$ instead of directly pouring at $\textbf{p}_{E}$ (Eqn. \ref{eqn:3}) for two reasons.
First, $\textbf{p}_{E}$ is obtained by freely dropping the particles, \textit{i.e.}, the particles have no horizontal velocities. 
Whereas in pouring, the particles have a horizontal velocity when exiting the cup mouth.
We indent the cup ``backward'' horizontally to compensate for this velocity.
Also, we want to explore more area for pouring. 
Indenting the cup towards the object periphery provides a wider exploration area and a higher chance to find a $(\theta_{pour}, \textbf{p}_{pour})$ with a large $P$ (Fig. \ref{fig:5}).

The $(\theta_{pour}^{*}, \textbf{p}_{pour}^{*}) = (\theta_{pour}(i^{*}), \textbf{p}_{pour}(i^{*}, j^{*}))$ for robot autonomous pouring is chosen by setting $i^*$ and $j^{*}$ as follows:
\begin{equation}
\begin{gathered}
    i^{*} = \argmax_{i} \sum_{j} P(\theta_{pour}(i), \textbf{p}_{pour}(i, j))\\
    j^{*} = \argmax_{j} P(\theta_{pour}(i^{*}), \textbf{p}_{pour}(i^{*}, j))
\end{gathered}
\end{equation}
We select $\theta_{pour}^{*}$ as above because pouring with this $\theta_{pour}(i^{*})$ has higher chances to get a large $P$, \textit{i.e.}, less spillage, in the real world.
The cup initial pose of pouring $(R_{init}, \textbf{t}_{init}) \in SE(3)$ in the world frame can be retrieved as follows:
\begin{equation}
\begin{gathered}
    R_{init} = R_{z}(\theta_{pour}^{*})R_{0}\\ 
    \textbf{t}_{init} = \begin{bmatrix}
        \textbf{p}_{pour}^{*} + \textbf{p}_{E}\\
        z_{E}
    \end{bmatrix} - R_{init}\textbf{r}^{'}_{pivot}
\end{gathered}
\label{eqn:6}
\end{equation}
where $R_{z}(\theta_{pour}^{*})$ specifies the rotation which rotates about the z-axis of the world frame by $\theta_{pour}^{*}$; $R_{0}$ rotates the cup such that $l_{cup}$ is parallel to the x-axis of $E$ and $\textbf{r}_{pivot}$ is the lowest end of the cup mouth in the world frame; $\textbf{r}^{'}_{pivot} \in \mathbb{R}^{3}$ is the position of the pivot in the object body frame; $\textbf{p}_E$ and $z_E$ are defined at the beginning of this section.

\section{Experiments}
Fig. \ref{fig:1}(b) shows the experiment setup. We implement our method on a UR5 robot mounted with an afag EU-20 UR gripper. 
A PrimeSense Carmine 1.09 RGB-D camera is also mounted on the end effector. 
The robot base frame is used as the world frame throughout the experiment.

\subsection{Robot 3D Scanning}
\label{sec: 4A}
In the experiment, the object of interest is placed randomly on a transparent platform in its upright pose. 
The transparent platform is also randomly placed on a table within limits that are described below.
Since the depth sensor of the camera has a shortest range of 0.35m, although random, the position of the object needs to satisfy: 1) the object does not collide with the robot during the scanning; 2) the object falls within the range of the depth sensor in all capturing views (\textit{e.g.}, the front-to-back view shown in Fig. \ref{fig:6}(a)); 3) the object is in the robot workspace so that the robot can pour into the object if it is classified as an open container.
Using the wrist mounted RGB-D camera, we move the robot's end effector to 24 pre-defined configurations to capture depth images of the scene. 
From the robot's forward kinematics, we are able to obtain the pose of the camera at each view. 
This allows us to use TSDF Fusion \cite{curless1996volumetric} to densely reconstruct the scene. 
Since the object is placed on a transparent platform which will not be captured by the depth sensor of the camera, we are able to segment the object from the scene by simply cropping the 3D reconstruction with a box\footnote{Plane segmentation can segment the object from the table without the transparent platform. However, the bottom of the inner surface of some objects (\textit{e.g.}, the red bag in Fig. \ref{fig:4}(b)) would be very close to the table and falsely segmented out as part of the table due to sensor inaccuracy.}.
Finally, we get the 3D mesh of the object.
It is worth noting that object segmentation could also be achieved without the transparent platform by 3D object segmentation \cite{finman2013toward} or physically picking up the object for scanning \cite{krainin2011manipulator}.

As we are not able to capture views from underneath the object, the reconstructed 3D model is partial. It only covers surfaces which can be seen from the top and side views. 
The inner containing surface of open containers can be seen from the top and side views given the object is placed upright, and thus is included in the reconstructed model. 
Therefore, the partially reconstructed model is sufficient for the purpose of open containability reasoning. 
Reconstructing the complete model of an object can be achieved by methods involving manipulating the object such as \cite{krainin2011manipulator, welke2010autonomous}.

\subsection{Data}
\begin{figure*}
    \centering
    \includegraphics[width=2\columnwidth]{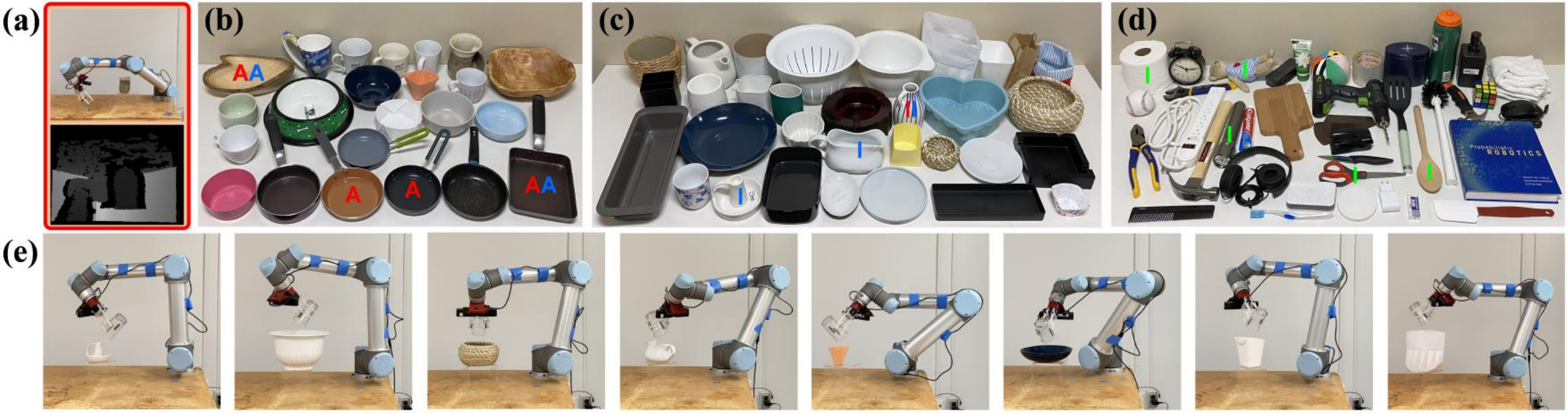}
    \caption{\textbf{Experiments and Results.} (a) shows one of the 24 views for 3D scanning. (b) and (c) together show all the open containers in the full test set where those in (b) are in the \textit{restricted} test set and those in (c) are the complement. The red and blue A's (I's) in (b) ((c)) show all the pouring failure cases of AffordanceNet (Imagination) and its variation, respectively. Note that Imagination has no pouring failure cases on the objects in the restricted test set (b); AffordanceNet is not tested on the objects in (c). For more information, see Sec. \ref{sec: 5}. (d) shows some examples of non-open containers in the full test set. The green I's show all the open container classification failure cases of Imagination on the full test set. (e) Pouring into 8 different open containers in the test set. There is no spillage during pouring into each of these containers. Notice how $\theta_{pour}$ varies for different objects.}
    \label{fig:6}
    \vspace{-0.5cm}
\end{figure*}

Our dataset contains 141 real-world objects which are commonly encountered in daily life (Fig. \ref{fig:6}(b)(c)(d)).
We use 11 objects for simulation calibration (calibration set) and the remaining 130, which are previously unseen by the robot, for testing (test set).
The 130 objects in the test set covers 57 object categories.
Since the objects have no labels of open container, we recruit 20 human subjects to annotate the data for open container classification. 
Every object is annotated 5 times by 5 different subjects.
Given an object, we asked each of the 5 subjects\footnote{We did not directly ask whether the object is an open container because the concept of open container is not clear to some human subjects.}: 1) Given the object in this pose, is it able to contain M\&M's\textsuperscript{\textregistered} candies? 2) Given the object in this pose, are you able to pour M\&M's\textsuperscript{\textregistered} candies into it?
If the annotations to these two questions are both true, we consider the object is annotated as an open container by the subject; otherwise, it is annotated as a non-open container.
If 3 or more subjects annotate an object as an open container, the ground truth label of the object is an open container; otherwise, the label is a non-open container.
The test set contains 55 open containers and 75 non-open containers.

\subsection{Physical Simulation}
\label{sec: 4c}
We use Pybullet \cite{coumanspybullet} as the physical engine for open containability imagination and pouring imagination. 
The object, cup, and particles are imported with the URDF files which specify the mass, CoM, inertia matrix, collision model, and friction coefficient.
The particles are modelled as ellipsoids with similar dimensions of an M\&M's\textsuperscript{\textregistered} candy.
We use Volumetric Hierarchical Approximate Convex Decomposition (V-HACD)\cite{mamou2016volumetric} to decompose the models of the object and the cup into convex pieces as the collision models. 
The default Coulomb friction model is used for friction modeling.
The collision between the particles and the object is modelled as inelastic (coefficient of restitution $e=0.1$).

We calibrate the simulation by running the algorithm on the calibration set.
The simulation parameters of the open containability imagination are manually calibrated such that the open container classification result matches with the human annotation.
The simulation parameters of the pouring imagination are manually calibrated such that the pouring result, \textit{i.e.}, $P$ in Eqn. \ref{eqn:pour}, matches with the real-world pouring result to the greatest extent.
From the simulation calibration, $\Omega>0$ for all the open containers in the calibration set; $\Omega=0$ for all the non-open containers.
Thus, we set $\Omega_{thr} = 0$ which physically means that an object is classified as an open container if there are particles retained within the object after the drop and a non-open container if otherwise.
We set $N_{pour}=60$ for all pourings in the pouring imagination, a number of which the volume can be contained by all the open containers in the dataset.
However, we want to point out that $N_{pour}$ can be set according to how much the user wants to pour and the containing volume of the object, which we did not cover in this paper.
A further investigation of reasoning $N_{pour}$ is left for future work.
The cup used in pouring imagination has the same dimensions as the one used in robot autonomous pouring (Sec. \ref{sec:4D}).

\subsection{Robot Autonomous Pouring}
\label{sec:4D}
In robot autonomous pouring, the robot pours the same number of candies as in pouring imagination.
At the beginning of each experiment, we fill the cup with candies and place it at a pre-defined position.
After the pouring imagination, the robot first picks up the cup and execute a pre-defined trajectory to move the cup to a pre-pour pose. 
The trajectory is designed such that the cup does not spill the candies during the execution. 
Then, the robot is controlled to move the cup to the imagined initial pose of pouring $(R_{init}, \textbf{t}_{init})$ obtained from the pouring imagination (Eqn. \ref{eqn:6}). 
After that, similar to the pouring imagination, the robot rotates the cup about $l_{pour}$ to pour the candies (Fig. \ref{fig:6}(e)).
We use the python-urx library \cite{pythonurx} to control and generate motion plans for the robot.

\section{Results}
\label{sec: 5}
In this section, we show the results of open container classification and robot autonomous pouring on the test set.
We evaluate our method on a computer running Intel Core i7-8700 @ 3.2GHz CPU. 
Our single-threaded unoptimized implementation takes about 50 seconds for 3D scanning, 15 seconds for open containability imagination, 80 seconds for pouring imagination, and 40 seconds for autonomous pouring.
We denote our method as \textbf{Imagination}.
We compare with a deep learning method \cite{do2018affordancenet}, denoted as \textbf{AffordanceNet}, which we find the most relevant in the literature.
AffordanceNet takes as input an RGB image.
It has an object detection branch which localizes and classifies objects with bounding boxes and an affordance detection branch which assigns each pixel within the bounding box with an affordance label.
AffordanceNet is trained on the IIT-AFF dataset \cite{nguyen2017object} which contains 8,835 real-world images with 10 object classes and 9 affordance classes.
Among the 10 object classes, 3 classes are open containers (cup, bowl, pan).
Among the 9 affordance classes, one of them is ``contain''.
Since AffordanceNet is not able to predict a label of object classes which does not belong to its training object classes, it is unfair to evaluate AffordanceNet's performance on objects which are not in the 10 object classes it is trained on. 
Therefore, the comparison with AffordanceNet is evaluated on a subset of the full test set, denoted as the \textit{restricted} test set.
It contains only 51 objects (23 open containers (Fig. \ref{fig:6}(b)) and 28 non-open containers) which falls in the object classes of the IIT-AFF dataset.
For each of these objects, at least four of the five human annotators were in agreement.

\subsection{Open Container Classification}
We use the classification accuracy (accuracy) and area under the Receiver Operating Characteristic curve (AUC) to evaluate the open container classification performance of different methods.
The results are shown in Table \ref{table:1}.
For our proposed method, the open containability $\Omega$ (Eqn. \ref{eqn:1}) is used as the confidence score to calculate the AUC.
For AffordanceNet, we use its object detection branch for open container classification.
Although our method only uses the depth images from the 24 views captured by the RGB-D camera, we also save the RGB images. 
These RGB images are used for testing AffordanceNet.
Our method segments the object (Sec. \ref{sec: 4A}) before the imagination for open container classification.
To ensure fairness to the greatest extent, we manually crop the object from each RGB image to subtract the background before inputting into AffordanceNet.
From the output of AffordanceNet's detection branch, we first select the detection (bounding box + classification score) with the highest classification score.
The maximum classification score of the 3 open container classes of this detection is used as the confidence score of the object being an open container.
We evaluate the classification accuracy and AUC with AffordanceNet for the 24 views, respectively.
The highest accuracy and AUC is selected as the accuracy and AUC of AffordanceNet.

\begin{table}[!htp]
\caption{Open Container Classification Results(\%)}
\centering
{
\begin{tabular}{c c c c}
\toprule
 Method & Dataset & Accuracy & AUC \\
 \arrayrulecolor{black}\midrule
  AffordanceNet & restricted & 100.00 & 100.00 \\
  Imagination   & restricted & 100.00 & 100.00  \\
 \arrayrulecolor{black!30}\midrule
  Imagination & full & \textbf{96.15} & \textbf{100.00}  \\
\arrayrulecolor{black}\bottomrule
\end{tabular}
}
\label{table:1}
\vspace{-0.2cm}
\end{table}

Both our method and AffordanceNet achieve perfect performance on the restricted test set.
On the full test set, our method fails to classify some objects (indicated with green I's in Fig.\ref{fig:6}(d)), labelled as non-open containers, the same way as the human annotation for two reasons.
Firstly, the reconstructed 3D model of the object traps particles with small concavities due to inaccurate reconstruction and approximate convex decomposition.
Secondly, the object (\textit{e.g.}, a spoon) possesses small concavities which can retain particles.
Interestingly, the spoon, classified as an open container by our method, is annotated as a non-open container by 3 out of 5 human subjects, \textit{i.e.}, whether it is an open container diverges among human subjects.

\subsection{Robot Autonomous Pouring}
\label{sec: 5B}
We use the method described in Sec. \ref{sec:3B} and \ref{sec:4D} to pour candies into the open containers in the test set.
We compare with AffordanceNet by using the affordance detection branch.
We derive $(\theta_{pour}, \textbf{p}_{pour})$ with the affordance detection and use the same method described in Sec. \ref{sec:4D} to pour.
The view with the highest AUC is used to test AffordanceNet\footnote{This view also has the highest classification accuracy.}. 
It is a top-down view which is similar to the view used for pouring in \cite{do2018affordancenet}.
As in the open container classification, we crop the object from the image to ensure fair comparison.
Using the same method described in \cite{do2018affordancenet}, we set $\textbf{p}_{pour}$ by calculating the centroid of all the pixels predicted as the ``contain'' affordance in 2D and project it to 3D with the corresponding depth image and camera intrinsic parameters.
However, directly pouring at the projected point will result in collision with the object if the point is on the object surface.
Therefore, we set the $z$ component of $\textbf{r}_{pivot}$ by adding a fixed offset $\Delta h$ to the $z$ component of the projected point.
Since AffordanceNet is not able to specify $\theta_{pour}$, we use a fixed $\theta_{pour}$ in the experiment.
We also compare with a variation of AffordanceNet, denoted as \textbf{AffordanceNet w/ 3D Scanning} which incorporates the robot 3D scanning module in our method to provide the $z$ component of $\textbf{r}_{pivot}$.
Similar to our method, the $z$ component is set such that $\textbf{r}_{pivot}$ is 1cm above the object AABB.
In addition, we compare to a variation of our method, denoted as \textbf{Imagination w/ fixed $\theta_{pour}$}, in which $\theta_{pour}$ is fixed and equals to that used for AffordanceNet. 
We want to study the importance of reasoning about pouring orientations.

\begin{table}[!htp]
\caption{Robot Autonomous Pouring Success Rate(\%)}
\centering
{
\begin{tabular}{c c  c }
\toprule
 Method & Dataset & Success Rate \\
 \arrayrulecolor{black}\midrule
 AffordanceNet & restricted & 82.61 \\
 AffordanceNet w/ 3D Scanning & restricted & 91.30 \\
 Imagination w/ fixed $\theta_{pour}$ & restricted & 100.00 \\
 Imagination & restricted & 100.00 \\
 \arrayrulecolor{black!30}\midrule
 Imagination w/ fixed $\theta_{pour}$ & full & 94.55 \\
 Imagination & full & \textbf{98.18} \\ 

\arrayrulecolor{black}\bottomrule
\end{tabular}
}
\label{table:3}
% \vspace{-0.3cm}
\end{table}

For AffordanceNet and its variation, if the network detects ``contain'' affordance for an object and pour without spillage, we consider the pouring a success and failure if otherwise.
For our method and its variation, if the open container imagination classifies the object as an open container and pour without spillage, we consider the pouring a success and failure if otherwise.
The results are shown in Table \ref{table:3}.
Failure cases of all methods are shown in Fig. \ref{fig:6}(b)(c).
Since AffordanceNet is not able to specify the $z$ component of $\textbf{r}_{pivot}$, it struggles with very shallow containers (\textit{e.g.}, pans) due to candies bouncing off the object.
With the robot 3D scanning to provide the object height, AffordanceNet w/ 3D Scanning is able to cope with this problem but still fails on some objects (\textit{e.g.}, the wooden bowl in Fig. \ref{fig:6}(b) with a similar color as the table) due to incorrect affordance detection.
Both Imagination and its variation perform perfectly on the restricted test set.
On the full test set, the variation fails on a gravy boat with an opening resembling a slim rectangle.
Without the reasoning of $\theta_{pour}$, it is not able to find the pouring orientation in which the cup mouth points along the length of the rectangle, providing larger tolerances for the randomness in the candy's pouring trajectory.
The only failure case for Imagination on the full test set is a vase with a small opening which was reconstructed as a dent with very limited depth due to limitations on the 3D scanning. The robot classifies it correctly but fails to pour without spillage.

\section{Discussion \& Future Work}
Despite being calibrated with 11 objects, our method is able to achieve the same open container classification performance as the deep learning method, which has been trained with thousands of annotated images, on the restricted test set. 
Moreover, our method's performance on the full test set aligns well with human judgements.
The pouring experiment results show that our method outperforms the deep learning method on the restricted test set and is able to maintain similar performance on the full test set.
Both the classification and pouring results of our method on the full test set, which covers 57 object categories, show that our method is able to achieve inter-class function (open containability) generalization which is challenging for appearance-based method for the reasons of data burden and the implicit nature of appearance representation.
We believe this generalization is a fundamental strength of the affordance-based object perception which physically captures the most essential cue of open containers through reasoning interactions with objects.

Unlike the black box nature of the deep learning method, our method is \textit{explainable}.
It explains the open containability affordance of an object by the number of particles the object is able to retain in the physical simulation.
Our method explores object affordances based on object instances instead of object categories.
For objects which do not belong to a typical open container categories but are able to afford open containability, \textit{e.g.}, the candlestick in Fig. \ref{fig:5}, our method is able to \textit{improvise} and recognize its affordance.
This provides robots with intelligence to tackle previously unseen object instances and/or categories when deployed in our daily life.

For future work, we plan to introduce a more accurate and adaptive scanning algorithm to tackle the failure cases in the experiments.
For more complicated real-world settings (\textit{e.g.}, cluttered scenes), we plan to introduce manipulation to physically segment objects from the scene by picking up the objects \cite{zeng2017multi, gualtieri2016high}. The objects can then be scanned to get their 3D models for imagination \cite{krainin2011manipulator}.

\section{Conclusion}
In this work, we develop a novel method for robots to imagine the open containability of previously unseen objects via physical simulations.
Our method is able to perform open container classification and endows robots with the capability to autonomous pour into objects via open containability imagination and pouring imagination, respectively.
We evaluate our method on a dataset containing 130 previously unseen objects with 57 object categories.
Results show that our method's performance on open container classification aligns well with human judgements and it is able to pour into the 55 open containers in the dataset with a success rate of 98.18\%.
This ability to access and generalize function to previously unseen object instances and object categories provides robots with high intelligence on robot-object interaction.
We hope that our method will serve as an effective approach to reinforce robot-object interaction in future research.

% Can use something like this to put references on a page
% by themselves when using endfloat and the captionsoff option.
\ifCLASSOPTIONcaptionsoff
  \newpage
\fi

\bibliographystyle{IEEEtran}
\bibliography{reference.bib}

\end{document}